\documentclass{article}

\usepackage{times}
\usepackage{graphicx} 
\usepackage{subfigure} 
\usepackage{natbib} 
\usepackage{algorithm} 
\usepackage{algorithmic}
\usepackage{amsfonts}
\usepackage{multirow}
\usepackage{amsmath}
\usepackage{xcolor}
\DeclareMathOperator{\argmax}{argmax}

\usepackage{hyperref}

\usepackage[accepted]{icml2017}

\icmltitlerunning{Sharper sensitivity maps: removing noise by adding noise}

\begin{document} 

\twocolumn[
\icmltitle{SmoothGrad: removing noise by adding noise}



\icmlsetsymbol{equal}{*}

\begin{icmlauthorlist}
\icmlauthor{Daniel Smilkov}{goog}
\icmlauthor{Nikhil Thorat}{goog}
\icmlauthor{Been Kim}{goog}
\icmlauthor{Fernanda Vi\'egas}{goog}
\icmlauthor{Martin Wattenberg}{goog}
\end{icmlauthorlist}

\icmlaffiliation{goog}{Google Inc.}

\icmlcorrespondingauthor{Daniel Smilkov}{smilkov@google.com}

\icmlkeywords{neural network, saliency mask, sensitivity map, visualization, debugging}

\vskip 0.3in
]



\printAffiliationsAndNotice{}

\begin{abstract} 

Explaining the output of a deep network remains a
challenge. In the case of an image classifier, one type of explanation is to identify pixels that strongly influence the final decision. A starting point for this strategy is the gradient of the class score function with respect to the input image. This gradient can be interpreted as a \textit{sensitivity map}, and there are several techniques that elaborate on this basic idea. This paper makes two contributions: it introduces \textsc{SmoothGrad}, a simple method that can help visually sharpen gradient-based sensitivity maps, and it discusses lessons in the visualization of these maps. We publish the code for our experiments and a website with our results.

\end{abstract} 

\section{Introduction}

Interpreting complex machine learning models, such as deep neural networks,
remains a challenge. 
Yet an understanding of how such models function is important both for building applications and as a problem in its own right. 
From health care domains~\cite{hughes2016supervised, doshi2014comorbidity,lou2012intelligible} to education~\cite{kim2015ibcm}, there are many domains where interpretability is important. For example, the pneumonia risk prediction case study in~\cite{lou2012intelligible} showed that more interpretable models can reveal important but surprising patterns in the data that complex models overlooked. For reviews of interpretable models, see \cite{freitas2014comprehensible, DoshiKim2017Interpretability}. 

One case of interest is image classification systems. Finding an ``explanation'' for a classification decision could potentially shed light on the underlying mechanisms of such systems, as well as helping in enhancing them. For example, the technique of deconvolution helped researchers identify neurons that failed to learn any meaningful features, knowledge that was used to improve the network, as in~\cite{zeiler2014visualizing}.

A common approach to understanding the decisions of image classification systems is to find regions of an image that were particularly influential to the final classification. \cite{baehrens2010explain, zeiler2014visualizing, springenberg2014striving, zhou2016learning, selvaraju2016grad, sundararajan2017axiomatic, welling2016new}.
These approaches (variously called sensitivity maps, saliency maps, or pixel attribution maps; see discussion in Section~\ref{sec:gradients}; use occlusion techniques or calculations with gradients to assign an ``importance'' value to individual pixels which is meant to reflect their influence on the final classification. 

In practice these techniques often do seem to highlight regions that can be meaningful to humans, such as the eyes in a face. At the same time, sensitivity maps are often visually noisy, highlighting some pixels that--to a human eye--seem randomly selected. Of course, a priori we cannot determine if this noise reflects an underlying truth about how networks perform classification, or is due to more superficial factors. Either way, it seems like a phenomenon worth investigating further.

This paper describes a very simple technique, \textsc{SmoothGrad}, that in practice tends to reduce visual noise, and also can be combined with other sensitivity map algorithms. The core idea is to take an image of interest, sample similar images by adding noise to the image, then take the average of the resulting sensitivity maps for each sampled image. We also find that the common regularization technique of adding noise at training time \cite{bishop1995training} has an additional ``de-noising'' effect on sensitivity maps. 
The two techniques (training with noise, and inferring with noise) seem to have additive effect; performing them together yields the best results. 

This paper compares the \textsc{SmoothGrad} method to several gradient-based sensitivity map methods and demonstrates its effects. We provide a conjecture, backed by some empirical evidence, for why the technique works, and why it might be more reflective of how the network is doing classification. We also discuss several ways to enhance visualizations of these sensitivity maps. Finally, we also make the code used to generate all the figures in this paper available, along with 200+ examples of each compared method on the web at \url{https://goo.gl/EfVzEE}.

\section{Gradients as sensitivity maps}
\label{sec:gradients}
Consider a system that classifies an image into one class from a set $C$. Given an input image $x$, many image classification networks~\cite{szegedy2016rethinking, lecun1998gradient} compute a class activation function $S_c$ for each class $c \in C$, and the final classification $class(x)$ is determined by which class has the highest score. That is,

$$class(x) = \argmax_{c \in C} S_c(x)$$

A mathematically clean way of locating ``important'' pixels in the input image has been proposed by several authors, e.g., \cite{baehrens2010explain, simonyan2013deep, erhan2009visualizing}.
If the functions $S_c$ are piecewise differentiable, for any image $x$ one can construct a \textit{sensitivity map} $M_c(x)$  simply by differentiating $M_c$ with respect to the input, $x$. In particular, we can define

$$M_c(x)  = \partial S_c(x) / \partial x$$

Here $\partial S_c$ represents the derivative (i.e. gradient) of $S_c$.
Intuitively speaking, $M_c$ represents
how much difference a tiny change in each pixel of $x$ would make to the classification score for class $c$.
As a result, one might hope that the resulting map $M_c$ would highlight key regions.

\begin{figure}[htbp]
\begin{center}
\centerline{\includegraphics[width=2.5in]{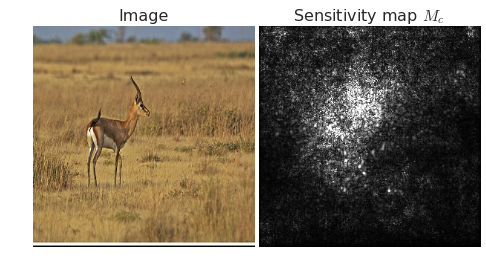}}
\caption{A noisy sensitivity map, based on the gradient of the class score for gazelle for an image classification network. Lighter pixels indicate partial derivatives with higher absolute values. See Section~\ref{sec:experiments} for details on the visualization.}
\label{noisy-grad}
\end{center}
\end{figure}

In practice, the sensitivity map of a label does seem to show a correlation with regions where that label is present~\cite{baehrens2010explain, simonyan2013deep}. However, the sensitivity maps based on raw gradients are typically visually noisy, as shown in Fig.~\ref{noisy-grad}. Moreover, as this image shows, the correlations with regions a human would pick out as meaningful are rough at best.

\subsection{Previous work on enhancing sensitivity maps}

 There are several hypotheses for the apparent noise in raw gradient visualizations. One possibility, of course, is that the maps are faithful descriptions of what the network is doing. Perhaps certain pixels scattered, seemingly at random, across the image are central to how the network is making a decision. On the other hand, it is also possible that using the raw gradient as a proxy for feature importance is not optimal. Seeking better explanations of network decisions, several prior works have proposed modifications to the basic technique of gradient sensitivity maps; we summarize a few key examples here.

One issue with using the gradient as a measure of influence is that an important feature may ``saturate'' the function $S_c$. In other words, it may have a strong effect globally, but with a small derivative locally. Several approaches, \textit{Layerwise Relevance Propagation} \cite{bach2015pixel},  \textit{DeepLift}~\cite{shrikumar2017learning}, and more recently \textit{Integrated Gradients} \cite{sundararajan2017axiomatic}, attempt to address this potential problem by estimating the global importance of each pixel, rather than local sensitivity. Maps created with these techniques are referred to as ``saliency'' or ``pixel attribution'' maps.

Another strategy for enhancing sensitivity maps has been to change or extend the backpropagation algorithm itself, with the goal of emphasizing positive contributions to the final outcome. Two examples are the \textit{Deconvolution}~\cite{zeiler2014visualizing} and \textit{Guided Backpropagation}~\cite{springenberg2014striving} techniques, which modify the gradients of ReLU functions by discarding negative values during the backpropagation calculation. The intention is to perform a type of ``deconvolution'' which will more clearly show features that triggered activations of high-level units. Similar ideas appear in \cite{selvaraju2016grad, zhou2016learning}, which suggest ways to combine gradients of units at multiple levels.

In what follows, we provide detailed comparisons of ``vanilla'' gradient maps with those created by integrated gradient methods and guided backpropagation. A note on terminology: although the terms ``sensitivity map'', ``saliency map'', and ``pixel attribution map'' have been used in different contexts, in this paper, we will refer to these methods collectively as ``sensitivity maps.''


\subsection{Smoothing noisy gradients}

There is a possible explanation for the noise in sensitivity maps, which to our knowledge has not been directly addressed in the literature: the derivative of the function $S_c$ may fluctuate sharply at small scales. In other words, the apparent noise one sees in a sensitivity map may be due to essentially meaningless local variations in partial derivatives. After all, given typical training techniques there is no reason to expect derivatives to vary smoothly. Indeed, the networks in question typically are based on ReLU activation functions, so $S_c$ generally will not even be continuously differentiable. 

\begin{figure}[htbp]
\begin{center}
\centerline{\includegraphics[width=3.2in]{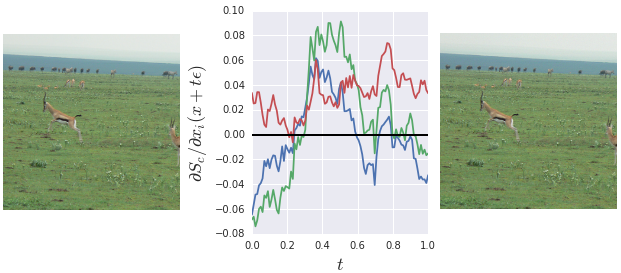}}
\caption{The partial derivative of $S_c$ with respect to the RGB values of a single pixel as a fraction of the maximum entry in the gradient vector, $\max_i \frac{\partial S_c}{\partial x_i}(t)$, (middle plot) as one slowly moves away from a baseline image $x$ (left plot) to a fixed location $x+\epsilon$ (right plot). $\epsilon$ is one random sample from $\mathcal{N}(0,\,0.01^2)$. The final image ($x + \epsilon$) is indistinguishable to a human from the origin image $x$.}
\label{noisy-der}
\end{center}
\end{figure}

Fig.~\ref{noisy-der} gives example of strongly fluctuating partial derivatives. This fixes a particular image $x$, and an image pixel $x_i$, and plots the values of $\frac{\partial S_c}{\partial x_i}(t)$ as fraction of  
the maximum entry in the gradient vector, $\max_i \frac{\partial S_c}{\partial x_i}(t)$,  for a short line segment $x+t\epsilon$ in the space of images parameterized by $t \in [0, 1]$. We show it as a fraction of the maximum entry in order to verify that the fluctuations are significant. The length of this segment is small enough that the starting image $x$ and the final image $x+\epsilon$ looks the same to a human. Furthermore, each image along the path is correctly classified by the model. The partial derivatives with respect to the red, green, and blue components, however, change significantly.


Given these rapid fluctuations, the gradient of $S_c$ at any given point will be less meaningful than a local average of gradient values. This suggests a new way to create improved sensitivity maps: instead of basing a visualization directly on the gradient $\partial S_c$, we could base it on a smoothing of $\partial S_c$ with a Gaussian kernel.

Directly computing such a local average in a high-dimensional input space is intractable, but we can compute a simple stochastic approximation. In particular, we can take random samples in a neighborhood of an input $x$, and average the resulting sensitivity maps. Mathematically, this means calculating

$$\hat{M_c}(x) = \frac{1}{n} \sum_1^n M_c(x + \mathcal{N}(0, \sigma^2))$$

where $n$ is the number of samples, and $\mathcal{N}(0, \sigma^2)$ represents Gaussian noise with standard deviation $\sigma$. We refer to this method as \textsc{SmoothGrad} throughout the paper.

\begin{figure*}
\begin{center}
\centerline{\includegraphics[width=6.5in]{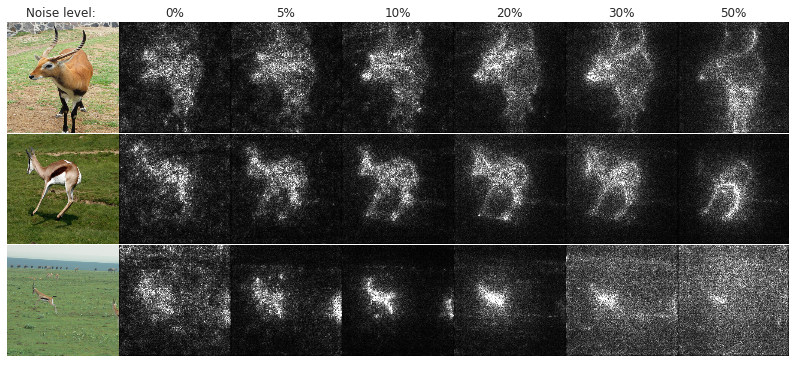}}
\caption{Effect of noise level (columns) on our method for 5 images of the gazelle class in ImageNet (rows). Each sensitivity map is obtained by applying Gaussian noise $\mathcal{N}(0,\,\sigma^{2})$ to the input pixels for $50$ samples, and averaging them. The noise level corresponds to $\sigma / (x_{max}-x_{min})$.}
\label{effect-noise-level-inception}
\end{center}
\end{figure*}

\section{Experiments}
\label{sec:experiments}
To assess the \textsc{SmoothGrad} technique, we performed a series of experiments using a neural network for image classification~\cite{szegedy2016rethinking, mnisttutorial}. The results suggest the estimated smoothed gradient, $\hat{M_c}$, leads to visually more coherent sensitivity maps than the unsmoothed gradient $M_c$, with the resulting visualizations aligning better--to the human eye--with meaningful features. 

Our experiments were carried out using an Inception v3 model \cite{szegedy2016rethinking} that was trained on the ILSVRC-2013 dataset \cite{russakovsky2015imagenet} and a convolutional MNIST model based on the TensorFlow tutorial \cite{mnisttutorial}.

\subsection{Visualization methods and techniques}

Sensitivity maps are typically visualized as heatmaps. Finding the right mapping from a channel values at a pixel to a particular color turns out to be surprisingly nuanced, and can have a large effect on the resulting impression of the visualization. This section summarizes some visualization techniques and lessons learned in the process of comparing various sensitivity map work. Some of these techniques may be universally useful regardless of the choice of sensitivity map methods.


\textbf{Absolute value of gradients}

Sensitivity map algorithms often produce signed values. There is considerable ambiguity in how to convert signed values to colors. A key choice is whether to represent positive and negative values differently, or to visualize the absolute value only.
The utility of taking the absolute values of gradients or not depends on the characteristics of the dataset of interest. For example, when the object of interest has the same color across the classes (e.g., digits are always white in MNIST digits \cite{lecun2010mnist}), the positive gradients indicate positive signal to the class. On the other hand, for ImageNet dataset \cite{russakovsky2015imagenet}, we have found that taking the absolute value of the gradient produced clearer pictures. One possible explanation for this phenomenon is that the direction is context dependent:
many image recognition tasks are invariant under color and illumination changes. For instance, in classifying a ball, a dark ball on a bright background would have negative gradient, while white ball on darker background would have a positive gradient.


\textbf{Capping outlying values}

Another property of the gradient that we observe is the presence of few pixels that have much higher gradients than the average. This is not a new discovery --- this property was utilized in generating adversarial examples that are indistinguishable to humans \cite{szegedy2013intriguing}. These outlying values have the potential to throw off color scales completely. Capping those extreme values to a relatively high value (we find $99^{\text{th}}$ percentile to be sufficient) leads to more visually coherent maps as in \cite{sundararajan2017axiomatic}. Without this post-processing step, maps may end up almost entirely black. 

\textbf{Multiplying maps with the input images}

Some techniques create a final sensitivity map by multiplying gradient-based values and actual pixel values \cite{shrikumar2017learning, sundararajan2017axiomatic}. 
This multiplication does tend to produce visually simpler and sharper images, although it can be unclear how much of this can be attributed to sharpness in the original image itself. For example, a black/white edge in the input can lead to an edge-like structure on the final visualization even if the underlying sensitivity map has no edges.

However, this may result in undesired side effect.
Pixels with values of $0$ will never show up on the sensitivity map. For example, if we encode black as $0$, the image of a classifier that correctly predicts a black ball on a white background will never highlight the black ball in the image. 

On the other hand, multiplying gradients with the input images makes sense when we view the importance of the feature as their contribution to the total score, $y$. For example, in a linear system $y=Wx$, it makes sense to consider $x_i w_i$ as the contribution of $x_i$ to the final score $y$.

For these reasons, we show our results with and without the image multiplication in Fig.~\ref{noisygrad-comparison}.

\subsection{Effect of noise level and sample size}

\textsc{SmoothGrad} has two hyper-parameters: $\sigma$, the noise level or standard deviation of the Gaussian perturbations, and $n$, the number of samples to average over. 

\textbf{Noise, $\sigma$}

\begin{figure*}[htbp]
\begin{center}
\centerline{\includegraphics[width=4in]{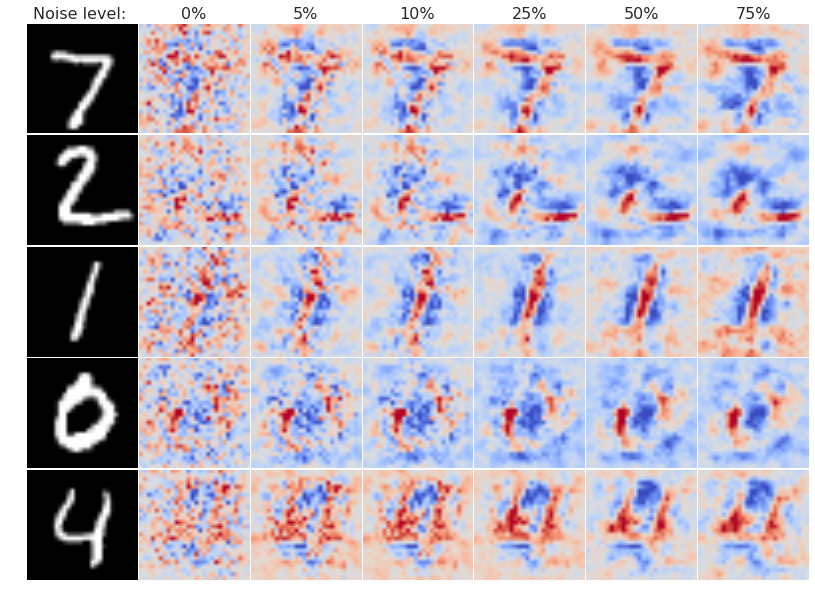}}
\caption{
Effect of noise level on the estimated gradient across 5 MNIST images. Each sensitivity map is obtained by applying a Gaussian noise at inference time and averaging in the same way as in Fig.~\ref{effect-noise-level-inception} over 100 samples.}
\label{effect-noise-level-mnist}
\end{center}
\end{figure*}

Fig.~\ref{effect-noise-level-inception} shows the effect of noise level for several example images from ImageNet \cite{russakovsky2015imagenet}. The $2^{\text{nd}}$ column corresponds to the standard gradient (0\% noise), which we will refer to as the ``Vanilla'' method throughout the paper. Since quantitative evaluation of a map remains an unsolved problem, we again focus on qualitative evaluation. We observe that applying 10\%-20\% noise (middle columns) seems to balance the sharpness of sensitivity map and maintain the structure of the original image.
We also observe that while this range of noise gives generally good results for Inception, the ideal noise level depends on the input. See Fig.~\ref{effect-noise-level-mnist} for a similar experiment on the MNIST dataset.

\textbf{Sample size, $n$}

\begin{figure*}
\begin{center}
\centerline{\includegraphics[width=6.5in]{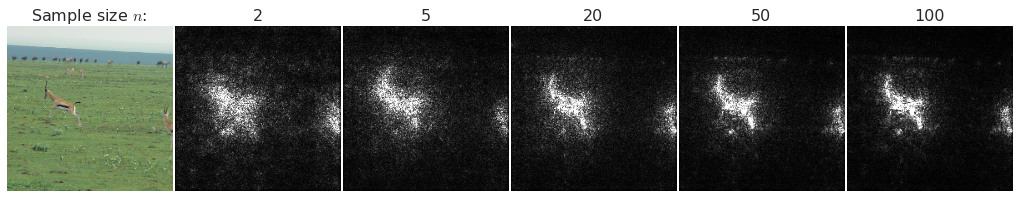}}
\caption{Effect of sample size on the estimated gradient for inception. 10\% noise was applied to each image.}
\label{effect-sample-size-inception}
\end{center}
\end{figure*}

In Fig.~\ref{effect-sample-size-inception} we show the effect of sample size, $n$. As expected, the estimated gradient becomes smoother as the sample size, $n$, increases. We empirically found a diminishing return --- there was little apparent change in the visualizations for $n > 50$.

\subsection{Qualitative comparison to baseline methods}



Since there is no ground truth to allow for quantitative evaluation of sensitivity maps, we follow prior work \cite{simonyan2013deep, zeiler2014visualizing, springenberg2014striving, selvaraju2016grad, sundararajan2017axiomatic} and focus on two aspects of qualitative evaluation.

First, we inspect visual coherence (e.g., the highlights are only on the object of interest, not the background). Second, we test for discriminativity, where in an image with both a monkey and a spoon, one would expect an explanation for a monkey classification to be concentrated on the monkey rather than the spoon, and vice versa.

\begin{figure*} 
\begin{center}
\centerline{\includegraphics[width=7.1in]{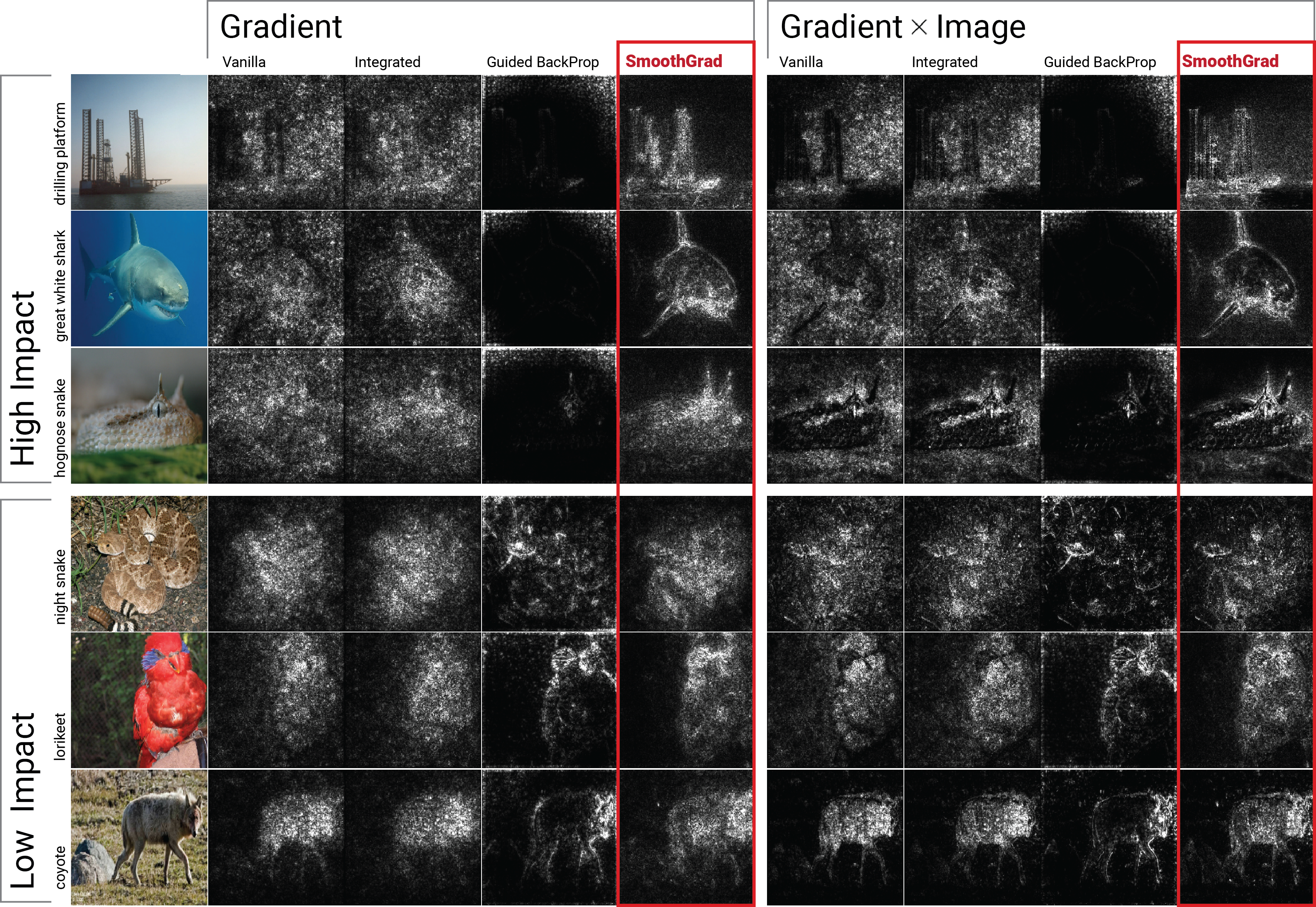}}
\caption{Qualitative evaluation of different methods. First three (last three) rows show examples where applying \textsc{SmoothGrad} had high (low) impact on the quality of sensitivity map.}
\label{noisygrad-comparison}
\end{center}
\end{figure*}

Regarding visual coherence, Fig.~\ref{noisygrad-comparison} shows a side-by-side comparison between our method and three gradient-based methods: \textit{Integrated Gradients} \cite{sundararajan2017axiomatic}, \textit{Guided BackProp} \cite{springenberg2014striving} and vanilla gradient. Among a random sample of $200$ images that we inspected, we found \textsc{SmoothGrad} to consistently provide more visually coherent maps than \textit{Integrated Gradients} and vanilla gradient. While \textit{Guided BackProp} provides the most sharp maps (last three rows of Fig.~\ref{noisygrad-comparison}), it is prone to failure (first three rows of Fig.~\ref{noisygrad-comparison}), especially for images with uniform background. On the contrary, our observation is that \textsc{SmoothGrad} has the highest impact when the object is surrounded with uniform background color (first three rows of Fig.~\ref{noisygrad-comparison}). Exploring this difference is an interesting area for investigation. It is possible that the smoothness of the class score function may be related to spatial statistics of the underlying image; noise may have a differential effect on the sensitivity to different textures.

\begin{figure*}[htbp]
\begin{center}
\centerline{\includegraphics[width=6.5in]{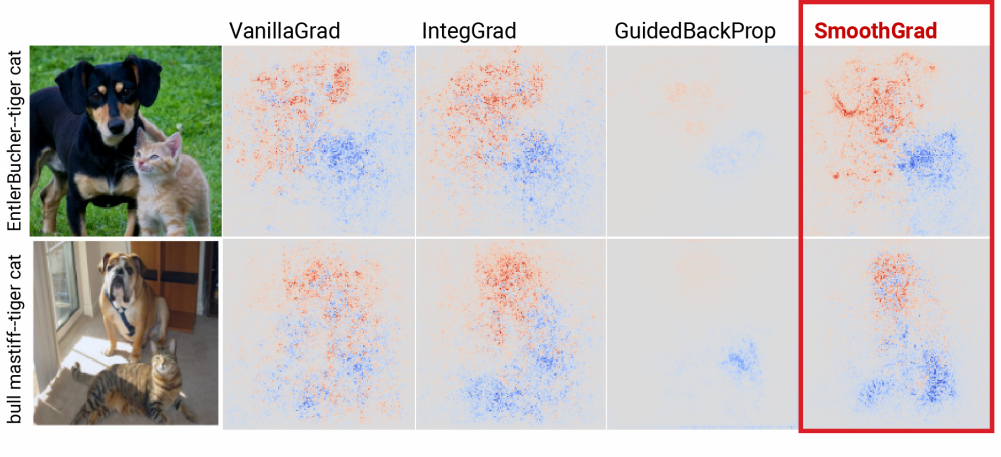}}
\caption{Discriminativity of different methods. For each image, we visualize the difference $\mbox{scale(} \partial y_1 / \partial x \mbox{)} - \mbox{scale(} \partial y_2 / \partial x \mbox{)}$ where $y_1$ and $y_2$ are the logits for the first and the second class (i.e., cat or dog) and scale() normalizes the gradient values to be between $[0, 1]$. The values are plotted using a diverging color map $[-1, 0, 1] \mapsto [\text{blue}, \text{gray}, \text{red}]$. Each method is represented in columns.}
\label{discriminative}
\end{center}
\end{figure*}

Fig.~\ref{discriminative} compares the discriminativity of our method to other methods. Each image has at least two objects of different classes that the network may recognize. To visually show discriminativity, we compute the sensitivity maps $M_1(x)$ and $M_2(x)$ for both classes, scale both to $[0, 1]$, and calculate the difference $M_1(x)-M_2(x)$. We then plot the values on a diverging color map $[-1, 0, 1] \mapsto [\text{blue}, \text{gray}, \text{red}]$.
For these images, \textsc{SmoothGrad} qualitatively shows better discriminativity over the other methods. It remains an open question to understand which properties affect the discriminativity of a given method -- e.g. understanding why \textit{Guided BackProp} seems to show the weakest discriminativity. 

\subsection{Combining SmoothGrad with other methods}

\begin{figure*}[htbp]
\begin{center}
\centerline{\includegraphics[width=4.5in]{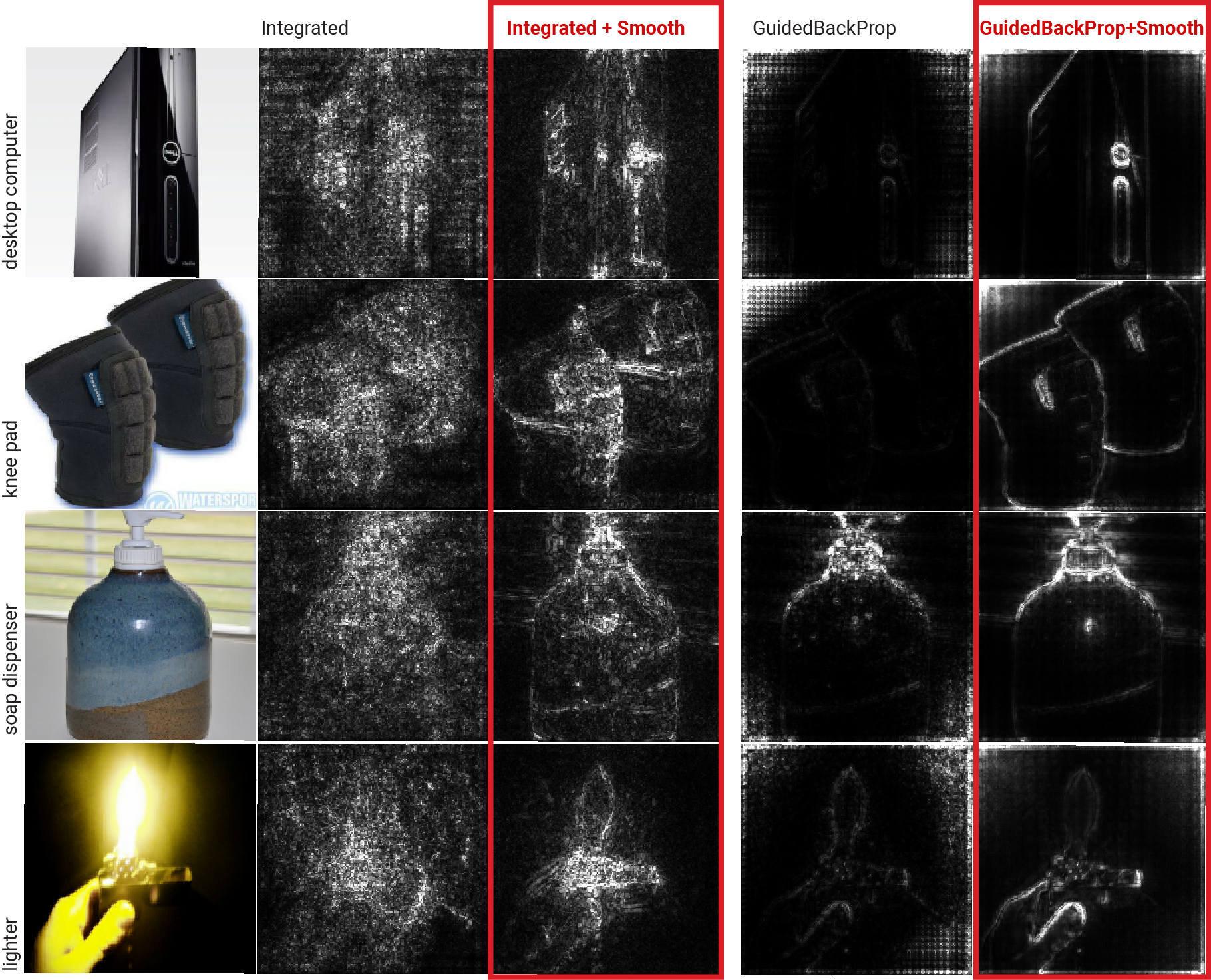}}
\caption{Using \textsc{SmoothGrad} in addition to existing gradient-based methods: \textit{Integrated Gradients} and \textit{Guided BackProp}.}
\label{augmented-methods}
\end{center}
\end{figure*}

One can think of \textsc{SmoothGrad} as smoothing the vanilla gradient method using a simple procedure: averaging the vanilla sensitivity maps of $n$ noisy images. With that in mind, the same smoothing procedure can be used to augment any gradient-based method. In Fig.~\ref{augmented-methods} we show the results of applying \textsc{SmoothGrad} in combination with \textit{Integrated Gradients} and \textit{Guided BackProp}. We observe that this augmentation improves the visual coherence of sensitivity maps for both methods.

For further analysis, 
we point the reader to our web page at \url{https://goo.gl/EfVzEE} with sensitivity maps of 200+ images and four different methods.

\subsection{Adding noise during training}

\textsc{SmoothGrad} as discussed so far may be 
applied to classification networks as-is.
In situations where there is a premium on legibility, however, it is natural to ask whether there is a similar way to modify the network weights so that its sensitivity maps are sharper. One idea that is parallel in some ways to \textsc{SmoothGrad} is the well-known regularization technique of adding noise to samples during training \cite{bishop1995training}. We find that the same method also improves the sharpness of the sensitivity map. 

\begin{figure}[htbp]
\begin{center}
\includegraphics[width=3.1in]{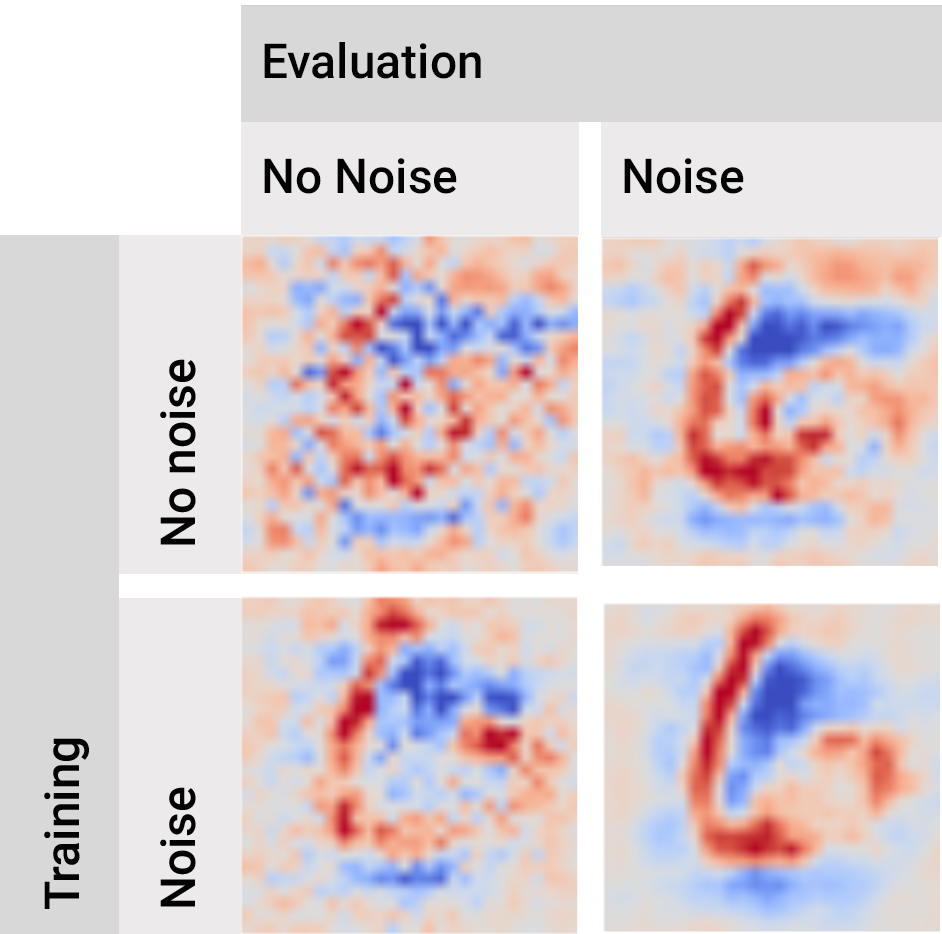}
\caption{Effect of adding noise during training vs evaluation for MNIST.}
\label{effect-train-eval-noise-mnist}
\end{center}
\end{figure}
\begin{figure}[htbp]
\begin{center}
\includegraphics[width=3.1in]{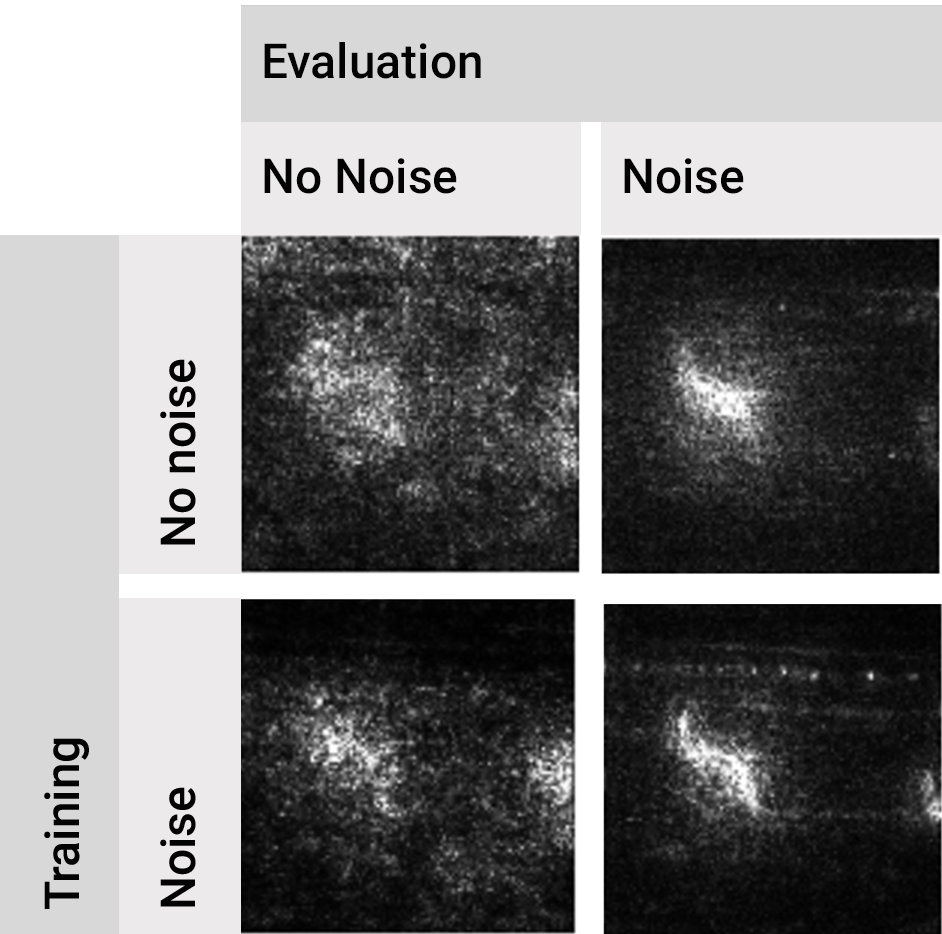}
\caption{Effect of adding noise during training vs evaluation for Inception.}
\label{effect-train-eval-noise-inception}
\end{center}
\end{figure}

Fig.~\ref{effect-train-eval-noise-mnist} and Fig.~\ref{effect-train-eval-noise-inception} show the effect of adding noise at training time and/or evaluation time for the MNIST and Inception model respectively. Interestingly, adding noise at training time seems to also provide a de-noising effect on the sensitivity map. Lastly, the two techniques (training with noise, and inferring with noise) seem to have additive effect; performing them together produces the most visually coherent map of the 4 combinations.

\section{Conclusion and future work}


The experiments described here suggest that gradient-based sensitivity maps can be sharpened by two forms of smoothing. First, averaging maps made from many small perturbations of a given image seems to have a significant smoothing effect. Second, that effect can be enhanced further by training on data that has been perturbed with random noise. 

These results suggest several avenues for future research. First, while we have provided a plausibility argument for our conjecture that noisy sensitivity maps are due to noisy gradients, it would be worthwhile to look for further evidence and theoretical arguments that support or disconfirm this hypothesis. It is certainly possible that the sharpening effect of \textsc{SmoothGrad} has other causes, such as a differential effect of random noise on different textures.

Second, in addition to training with noise, there may be more direct methods to create systems with smoother class score functions. For example, one could train with an explicit penalty on the size of partial derivatives. To create more spatial coherent maps, one could add a penalty for large differences in partial derivatives of the class score with respect to neighboring pixels. It may also be worth investigating the geometry of the class score function to understand why smoothing seems to be more effective on images with large regions of near-constant pixel values.

A further area for exploration is to find better metrics for comparing sensitivity maps. To measure spatial coherence, one might use existing databases of image segmentations, and we are already making progress \cite{oh2017exploiting, selvaraju2016grad}. Systematic measurements of discriminativity could also be valuable. Finally, a natural question is whether the de-noising techniques described here generalize to other network architectures and tasks.

\section*{Acknowledgements} 
We thank Chris Olah for generously sharing his code and helpful discussions, including pointing out the relation to contractive autoencoders, and Mukund Sundararajan and Qiqi Yan for useful discussions.

\bibliography{references}
\bibliographystyle{icml2017}

\end{document}